\documentclass[letterpaper, 10 pt, conference]{ieeeconf}  

\IEEEoverridecommandlockouts                              

\usepackage{graphics} 
\usepackage{epsfig} 
\usepackage{times} 
\usepackage{amsmath} 
\usepackage{amssymb}  

\usepackage{cite}
\usepackage[breaklinks=true,bookmarks=true,colorlinks]{hyperref}
\usepackage[capitalize]{cleveref}
\usepackage[table]{xcolor}
\usepackage{booktabs}
\usepackage{multirow}
\usepackage{colortbl}

\newcommand{\ours}{AtomSkill}

\title{\LARGE \bf
Learning Semantic Atomic Skills for Multi-Task Robotic Manipulation
}

\author{
Yihang Zhu$^{1,2,*}$, Weiqing Wang$^{1,2,*}$, Shijie Wu$^{1}$, Ye Shi$^{1,2}$, Jingya Wang$^{1,\dagger}$ \\
$^1$ShanghaiTech University, Shanghai, China \quad $^2$InstAdapt\\
{\tt\footnotesize \{zhuyh2023, wangwq2023, wushj12023, shiye, wangjingya\}@shanghaitech.edu.cn} \\
}

\begin{document}

\maketitle
\def\thefootnote{$*$}\footnotetext{Equal contributions.}
\def\thefootnote{$\dagger$}\footnotetext{Corresponding authors.}

\thispagestyle{empty}
\pagestyle{empty}

\begin{abstract}
Scaling imitation learning to diverse multi-task robot manipulation remains challenging due to suboptimal demonstrations, behavioral multi-modality, and destructive interference across tasks. While skill-based methods offer a promising direction by decomposing behaviors into reusable abstractions, existing approaches often learn skills that are either biased toward linguistic structure or lack semantic alignment across tasks, limiting generalization. In this work, we propose AtomSkill, a novel framework that learns a semantically aligned Atomic Skill Space from demonstrations and enables robust long-horizon execution through keypose imagination. Our method introduces: (1) semantic contrastive skill alignment, which partitions demonstrations into variable-length atomic skills and employs a contrastive objective to jointly enforce semantic consistency and temporal coherence, yielding a compact and reusable skill library; and (2) action decoding with keypose imagining, where the policy predicts both a skill’s terminal keypose and immediate actions, thereby supporting progress-aware skill transitions. During inference, an atomic skill diffusion sampler generates plausible skill sequences, while predicted keyposes autonomously trigger smooth skill chaining. Extensive experiments in simulation and real-world settings show that AtomSkill consistently outperforms state-of-the-art imitation learning and skill-based baselines.
Project page: \url{https://atom-skill.github.io}.
\end{abstract}
\section{Introduction}
\label{sec:intro}

Recent years have witnessed significant progress in imitation learning (IL) for robot manipulation. Methods such as DP~\cite{chi2023diffusion}, ACT~\cite{zhao2023learning}, and DP3~\cite{ze20243d} have demonstrated the ability to learn effective control policies directly from human teleoperated demonstrations. While these approaches excel in single-task settings, IL inherently struggles with fundamental challenges, including suboptimal expert demonstrations, trajectory noise, and behavioral multi-modality~\cite{o2024open,urain2024deep,zheng2022imitation}. These issues become critically amplified in multi-task scenarios, where increased data diversity and complexity often lead to policies that fail to generalize or suffer from destructive interference across tasks. Consequently, scaling imitation learning to a broad spectrum of tasks remains a fundamental and unresolved challenge. 

An intuitive approach for multi-task imitation learning is to structure manipulation behaviors by decomposing complex actions into reusable skill abstractions. This line of work can be broadly categorized into two complementary strands. One focuses on learning compressed latent representations of skills that can be shared across tasks. 
For instance, methods such as LISA\cite{garg2022lisa}, LCSD\cite{ju2024rethinking}, and SkillDiffuser~\cite{liang2024skilldiffuser} learn linguistic abstractions from language and visual input, without a smooth alignment with the action space, thus struggling to disentangle task-specific behaviors and model multimodal actions~\cite{wang2024sparse}.
Another branch of action-only methods learns skills directly from raw action sequences: Discrete Policy\cite{wu2025discrete} employs a VQ-VAE, while QueST\cite{mete2024quest} and STAR\cite{li2025star} enhance vector quantization with techniques such as finite scalar quantization~\cite{mentzer2023finite} and rotation tricks~\cite{fifty2024restructuring}. Despite their representation efficiency, their learned abstractions are primarily driven by low-level motion regularities, without explicitly modeling semantic equivalence between motion segments originating from different tasks. As a result, motions that share similar intent but appear in distinct task contexts are often encoded as separate or weakly related latent tokens, preventing effective cross-task alignment.
This lack of semantic alignment causes the learned skill space to emphasize local motion patterns rather than reusable, task-agnostic skills, ultimately constraining generalization in multi-task settings.

To address these challenges, we propose \ours{}, which learns a semantically aligned Atomic Skill Space across tasks and performs inference via keypose imagining. By aligning semantically equivalent motions across task contexts, AtomSkill builds a compact and reusable skill library. During execution, predicted terminal keyposes provide an explicit goal signal that supports progress-aware skill transitions.

Our approach is built around two core innovations. First, we introduce Atomic Skill Learning with Semantic Contrastive Skill Alignment, which segments demonstrations into variable-length atomic skills and employs a contrastive objective to jointly ensure both semantic consistency and temporal coherence in the learned skill embeddings. Our framework learns a VQ-VAE-style skill prior regularized by a contrastive objective, forming a compact and semantically grounded codebook. Second, our Action Decoder with Keypose Imagining jointly predicts both the skill’s terminal keypose and the immediate action sequence, enabling the policy to reason about long-horizon motion intent and fine-grained control simultaneously. Here, we introduce a novel inference paradigm driven by an Atomic Skill Diffusion Sampler, which generates plausible skill sequences from the library for robust high-level planning. This sampler is coupled with Action Chunking with Keypose, where the predicted keypose acts as a progress monitor to autonomously trigger smooth skill transitions, enabling reliable long-horizon execution without manual heuristics. Extensive experiments in simulation and real-world environments demonstrate that \ours\ consistently outperforms state-of-the-art imitation learning and skill-based baselines. Ablation studies confirm the contribution of each component, highlighting the importance of semantic-temporal alignment and keypose-conditioned chunking. 

In summary, our contributions are as follows:
\begin{itemize}
\item \textbf{AtomSkill Framework for Multi-Task Robot Manipulation.} We introduce AtomSkill, a novel multi-task imitation learning framework that learns and leverages a structured Atomic Skill Space to enable multi-task robot manipulation. 

\item \textbf{Semantically Grounded Atomic Skill Library.} We develop a skill learning paradigm that partitions demonstrations into variable-length atomic skills using gripper-state keyframe detection and vision-language model annotation. A contrastive learning objective ensures both semantic consistency and temporal coherence, creating a compact and reusable skill codebook.

\item \textbf{Action Generation with Keypose Imagination.} We design an action decoder that jointly predicts terminal keyposes (long-horizon intent) and immediate actions. This keypose imagination mechanism enables simultaneous reasoning about motion goals and fine-grained control, facilitating robust skill chaining and spatial understanding. 
\end{itemize}

\section{Related Work}
\label{sec:related_work}


\subsection{Multi-Task Imitation Learning}

Imitation learning (IL) enables robots to acquire skills by mimicking expert demonstrations~\cite{chi2023diffusion, zhao2023learning, ze20243d}. Early IL methods such as ACT~\cite{zhao2023learning}, Diffusion Policy~\cite{chi2023diffusion}, and DP3~\cite{ze20243d} focus on single-task settings, modeling conditional action distributions from visual and state inputs using generative frameworks like VAEs and diffusion models.

While effective in isolated tasks, extending IL to multi-task scenarios remains a key challenge. Some approaches tackle this by learning shared perception modules across tasks~\cite{shridhar2023perceiver, ke20243d, ze2023gnfactor,tian2025pdfactor}, though they typically predict keyframes rather than continuous actions, limiting applicability in complicated tasks. \cite{liu2026foam} encourages policy network to predict future images as goal conditions. Others draw inspiration from Mixture-of-Experts~\cite{shazeer2017sparsely, jacobs1991adaptive} and employ modular architectures where each expert or module implicitly learns different skills~\cite{wang2024sparse, antotsiou2022modular, reuss2024efficient, liu2025flexible, hao2026abstracting, deng2026semantically}. 
Beyond architectural designs, a growing line of work leverages latent variable models (LVMs) to uncover structured skill spaces that facilitate knowledge sharing across tasks and improve data efficiency.

\subsection{Latent Variable Models for Imitation Learning}
Latent variable models (LVMs) have been widely used in imitation learning to extract structured representations from complex demonstrations~\cite{liang2024skilldiffuser, garg2022lisa, lee2024behavior}. 
LISA~\cite{garg2022lisa}, LCSD~\cite{ju2024rethinking}, and SkillDiffuser~\cite{liang2024skilldiffuser} infer linguistic abstractions from language and visual input, providing high-level semantic representations, but they struggle to disentangle task-specific behaviors and model multimodal action realizations, often leading to policy conflation across distinct behaviors~\cite{wang2024sparse}. 
Recent works such as VQ-BeT~\cite{lee2024behavior}, Discrete Policy~\cite{wu2025discrete}, QueST~\cite{mete2024quest}, and ARP~\cite{wang2026arp} apply (Residual) VQ-VAE to learn discrete action abstractions, yet they are prone to codebook collapse or limited skill diversity~\cite{li2025star} due to representational bottlenecks and coarse temporal segmentation. 
STAR~\cite{li2025star} mitigates codebook collapse via rotation-based gradient alignment and adopts residual quantization for more expressive skills; however, similar to prior VQ-based methods, skill discovery is performed over overlapping motion segments, resulting in ambiguous skill boundaries and entangled semantics. 
In contrast, \ours{} discovers atomic skills from semantically decomposed action trajectories and leverages the open-world understanding ability of large vision-language models (VLMs) to explicitly ground skill boundaries in semantic structure.



\section{Method}
\label{sec:method}

\begin{figure*}[t]
    \centering
    \includegraphics[width=\linewidth]{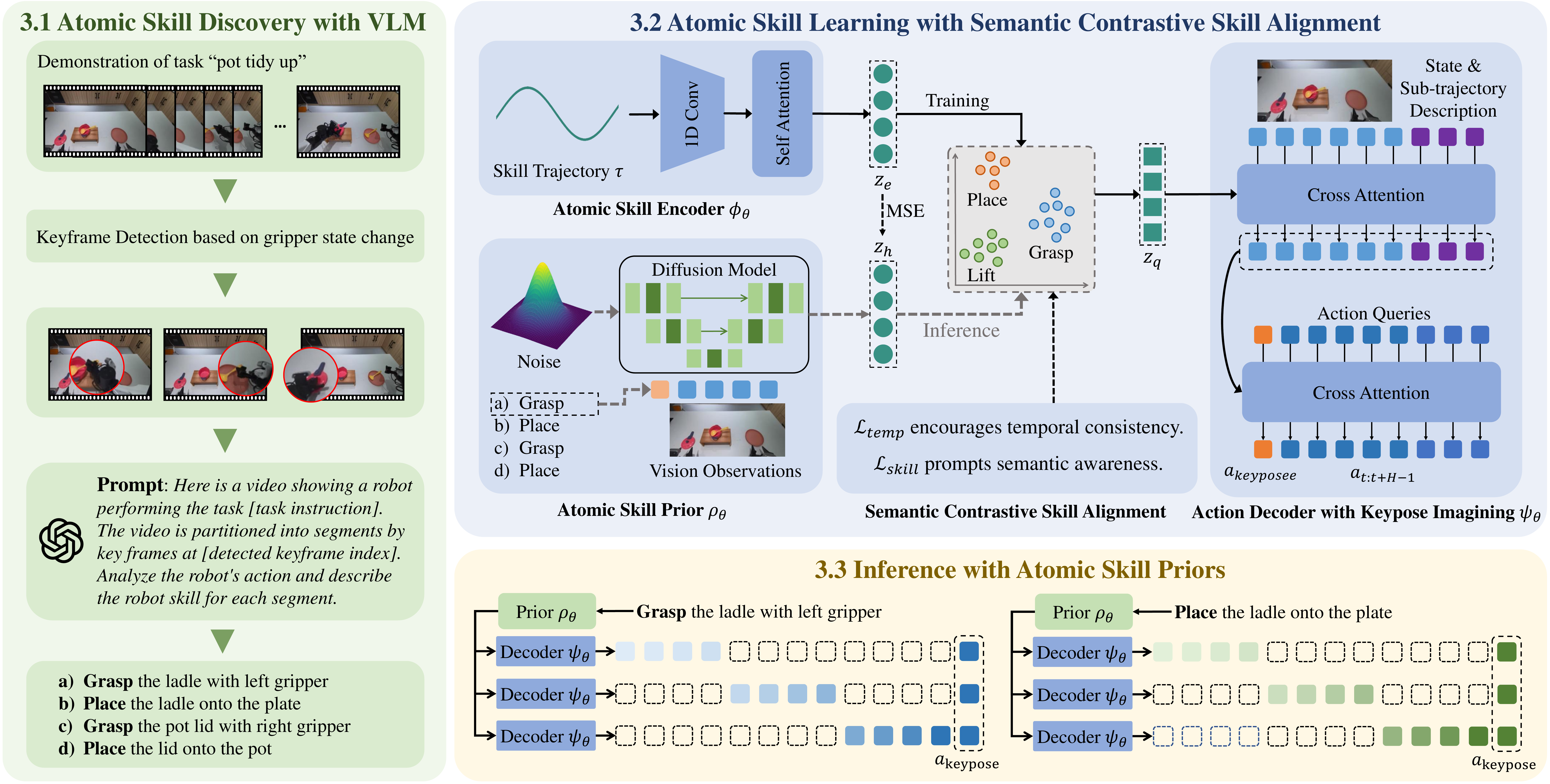}
    \caption{Framework of \ours. The left panel illustrates semantic skill discovery: expert demonstrations of the same task are segmented into semantically coherent, temporally aligned clips, and a vision–language model assigns a skill label to each segment. The top-right panel shows skill learning, where \ours\ structures the skill space and trains both the skill-guided policy and the diffusion-based sampler. The bottom-right panel depicts inference via action chunking with keypose, enabling smooth and robust chaining of predicted skills.}
    \label{fig: main figure}
\end{figure*}

In this section, we present \ours, a semantic skill-based imitation learning framework that learns semantically meaningful skill priors to provide high-level guidance across diverse tasks. An overview of the framework is shown in \cref{fig: main figure}. \ours\ comprises three main components:
(1) \textbf{Semantic Skill Discovery with VLM (\cref{subsec: skill discovery})}, which segments demonstrations into skill-centric clips that are both semantically coherent and temporally aligned, and leverages a large vision–language model to assign a skill label to each segment;
(2) \textbf{Atomic Skill Learning with Contrastive Alignment (\cref{subsec: skill vqvae})}, which structures the latent skill space by aligning it with open-world knowledge, while jointly training an action decoder with keypose imagining;
(3) \textbf{Inference with Skill Priors (\cref{subsec: inference})}, which combines a diffusion-based skill sampler for inference-time sampling, and an inference strategy that converts predicted skills into smooth and timely action sequences by chaining skills at keyframes.

\subsection{Semantic Skill Discovery with VLM} 
\label{subsec: skill discovery}

We define a skill as a set of action sequences that fulfil semantically identical objectives, such as "grasp" and "place", which is task-agnostic. Given an expert demonstration trajectory \(\tau\):
\begin{equation}
\tau = \left\{\{(O_t, a_t)\}_{t=1}^{T}, L\right\},
\end{equation}
where \(O_t\) and \(a_t\) denote the observation and action at timestep \(t\), \(L\) is a natural language instruction describing the task carried out in the demonstration, and we assume that the trajectory \(\tau\) is composed of a chain of skills.
The objective of semantic skill discovery is to split the trajectory into non-overlapping sub-trajectories \(\{\tau_1,\tau_2,...,\tau_n\}\) and obtain the corresponding skill label \(\{s_1,s_2,...,s_n\}\).
Following \cite{chen2025deco}, we realize such segmentation based on keyframes containing the change of gripper state. Despite its simplicity, gripper state change is a robust indicator of critical physical interaction, which can be viewed as a signal of the end of a skill and the start of the next skill.

To determine the corresponding skill label, a large vision-language model ~\cite{achiam2023gpt, bai2025qwen2} is employed to generate semantically grounded descriptions \(L_s\) and corresponding skill labels \(s\) for each segment, based on observations and task instructions. 
For example, given the task instruction ``put rubbish in bin'' the large VLM  produces sub-trajectory descriptions such as ``grasp the rubbish'' and ``place the rubbish in the bin'' which correspond to the atomic skill labels ``grasp'' and ``place'' respectively.
These labeled sub-trajectories \(\{(\tau_i, L_{s_i}, s_i)\}\) pave the way to learning modular and reusable skill representation.

\subsection{Atomic Skill Learning with Semantic Alignment}
\label{subsec: skill vqvae}

To tackle the challenge of discovering semantically meaningful skills from raw low-level action streams, we adopt a latent-variable formulation with an information bottleneck that extracts compact yet informative abstractions from control sequences. Specifically, we employ a VQ-VAE consisting of an encoder $\phi_{\theta}$, which compresses sub-trajectory actions into a fixed-length latent sequence, and a decoder $\psi_{\theta}$, which predicts future action sequences conditioned on both the latents and observations. Once the skill latent space has been structured, we further train a diffusion-based sampler $\rho_{\theta}$ to generate high-level skill embeddings.


\noindent\textbf{Skill Quantization.}
The skill encoder $\phi_\theta$ is implemented with several 1D CNN layers followed by self-attention layers.
Given a sub-trajectory $\tau_i$ obtained from \cref{subsec: skill discovery}, we extract the action sequence and resample it to a fixed size to handle variable durations.
The encoder $\phi_\theta$ processes this sequence to produce continuous embeddings $z_e$. 
A Vector Quantization (VQ) layer then maps $z_e$ to the nearest entries in a learnable codebook $\mathcal{E}=\{e_k\}_{k=1}^{K}$, yielding discrete skill tokens $z_q$. This quantization effectively discretizes the continuous motion space into atomic skill patterns.

\noindent \textbf{Semantic Contrastive Skill Alignment.}
Vector quantization (VQ) converts continuous action embeddings into discrete tokens, which capture shared motion patterns but inevitably discard fine-grained details.
Instead of treating this abstraction as a defect to be corrected~\cite{mete2024quest, li2025star, lee2024behavior}, we exploit it to learn high-level, semantically meaningful skill representations.

Given skill representations $z=z_e$, we employ supervised contrastive learning~\cite{khosla2020supervised} to align the latent space with semantic labels derived from \cref{subsec: skill discovery}. 
We formulate a general contrastive loss. Let the normalized similarity probability between an anchor $z_i$ and a candidate $z_p$ be:

\begin{equation}
    P(i, p) = \frac{\exp(z_i \cdot z_p / \tau)}{\sum_{z_a \in \mathcal{A}(i)} \exp(z_i \cdot z_a / \tau)},
\end{equation}

where $\tau$ is a temperature hyperparameter, $\mathcal{A}(i)$ denotes the set of all candidates contrasted against the anchor $z_i$.

The general contrastive objective is defined as:

\begin{equation}
    \mathcal{L}_{\text{SC}}(z_i, \mathcal{P}(i)) = - \frac{1}{|\mathcal{P}(i)|} \sum_{z_p \in \mathcal{P}(i)} \log P(i, p).
\end{equation}

Here, $\mathcal{P}(i)$ denotes the positive set for anchor $i$, and $|\mathcal{P}(i)|$ is its cardinality.

Based on this, we design two objectives to shape the latent space:
\begin{equation}
    \mathcal{L}_{\text{contrast}} = \sum_{i} \left( \mathcal{L}_{\text{SC}}(z_i, \mathcal{P}_{\text{temp}}(i)) + \mathcal{L}_{\text{SC}}(z_i, \mathcal{P}_{\text{skill}}(i)) \right).
\end{equation}
The first term, $\mathcal{L}_{\text{SC}}(\cdot, \mathcal{P}_{\text{temp}})$, encourages \textit{temporal consistency}. Here, $\mathcal{P}_{\text{temp}}(i)$ includes skill abstractions sharing the same quantized token $z_q$ and relative position within the skill sequence, ensuring structural coherence.
The second term, $\mathcal{L}_{\text{SC}}(\cdot, \mathcal{P}_{\text{skill}})$, promotes \textit{semantic awareness}. Here, $\mathcal{P}_{\text{skill}}(i)$ gathers abstractions sharing the same semantic skill label $s_k \in \mathcal{S}$ and quantized token, clustering functionally similar skills.
Finally, our total objective combines temporal and semantic alignment:

\noindent \textbf{Action Decoder with Keypose Imagining.}
At each timestep $t$, the action decoder $\psi_\theta$ receives the observation
\begin{equation}
O_t = \{I_t, p_t, L_s\},
\end{equation}
where $I_t = \{I_t^{\text{front}}, I_t^{\text{wrist}}\}$ denotes multi-view visual inputs, $p_t$ represents robot proprioception, and $L_s$ is the language instruction describing the current sub-task.
In addition, the skill sampler provides the skill abstraction $z_q$ corresponding to the current sub-trajectory $\tau_i$ (~\cref{fig: main figure}, right).

To inject skill information into the observation stream, we employ a cross-attention based action decoder $\psi_\theta$.
The cross-attention module treats the skill abstractions as keys and values, while the observation components $I_t$, $p_t$, and $L_s$ serve as the queries.
Compared with self-attention over the concatenated tokens, this design both reduces computational cost and explicitly amplifies the influence of skill latents on action generation.
Following ~\cite{zhao2023learning}, the decoder $\psi_\theta$ further performs cross-attention between fixed sinusoidal positional encodings and the fused observation features to model temporal dependencies.
Given the skill abstraction $z_q$ of a sub-trajectory $\tau_i$ and the current observation $O_t$, the action decoder predicts the future action chunk:
\begin{equation}
    \hat{a} = (\hat{a}_t, \ldots, \hat{a}_{t+H-1}) = \psi_\theta(O_t, z_q),
    \label{eq:chunk}
\end{equation}
with reconstruction loss $\mathcal{L}_a$:

\begin{equation}
    \mathcal{L}_a = \lVert a - \hat{a} \rVert_1.
\end{equation}

Thus, the total loss $\mathcal{L}$ can be written as a sum of VQ-VAE loss $\mathcal{L}_{\text{VQ}}$, reconstruction loss $\mathcal{L}_a$ and the contrastive loss $\mathcal{L}_{\text{contrast}}$:
\begin{equation}
    \mathcal{L} = \mathcal{L}_{\text{VQ}} + \beta_1\mathcal{L}_a + \beta_2\mathcal{L}_{\text{contrast}},
\end{equation}
where \(\beta_1\) and \(\beta_2\) are coefficients to balance different loss term.


\subsection{Inference with Atomic Skill Priors}
\label{subsec: inference}

\noindent \textbf{Atomic Skill Prior.} We model the skill prior $\rho_\theta$ as a diffusion model that samples from the learned skill space.
Diffusion models view data generation as an iterative denoising process and have been widely adopted in robotics for their expressiveness and flexibility~\cite{urain2024deep, chi2023diffusion, wu2025afforddp, liu2025sample}.
Starting from an initial noisy skill embedding $z^k$, the skill diffusion model uses a noise prediction network $\epsilon_\theta$ to iteratively remove noise conditioned on the skill label $s$ and observation $o$, until a clean skill embedding $z^0$ is obtained.
Similar to standard diffusion formulations~\cite{chi2023diffusion}, we train $\epsilon_\theta$ to predict the noise $\epsilon^k$ added at the $k$-th step:

\begin{equation}
    \mathcal{L}_{\text{sampler}} = \lVert \epsilon_k - \epsilon_\theta(z^k, k, o, s) \rVert^2.
    \label{eq:skill_prior}
\end{equation}

At inference time, the skill diffusion model starts from Gaussian noise and recovers a high-level skill embedding $z_h$ through $k$ denoising steps.
The resulting embedding is then mapped through the codebook and quantized into a discrete skill latent $z_q$, which is fed into the skill-guided policy and fused with the current observations to decode the corresponding action sequence.

\noindent \textbf{Action Chunking with Keypose.}
Our policy generates behavior by executing action chunks. To chain these skills effectively, the policy must both faithfully realize the current skill and reliably decide when to transition to the next one. To this end, we introduce action chunking with keypose imagining: in addition to predicting the subsequent action sequence, the decoder $\psi_\theta$ is trained to explicitly predict the next-keyframe action, which we refer to as the keypose $a_{\text{keypose}}$. This design offers two main advantages.

First, keypose prediction strengthens the policy’s spatial reasoning and localization ability. As illustrated at the bottom of ~\ref{fig: main figure}, the skill sampler $\rho_\theta$ is queried only when a new skill is required; it samples a high-level skill embedding that specifies the current skill to execute. Conditioned on the current observation and this embedding, the skill-guided policy $\psi_\theta$ rolls out a chunk of low-level actions and simultaneously predicts the keypose corresponding to the termination of the current skill. To obtain this keypose prediction, we append an additional query token to the action queries and reuse the same decoder to output $\hat{a}_{\text{keypose}}$, which is trained with an auxiliary loss:
\begin{equation}
    \mathcal{L}_{\text{keypose}} = \lVert a_{\text{keypose}} - \hat{a}_{\text{keypose}} \rVert_1 .
    \label{eq:keypose}
\end{equation}
By explicitly regressing the terminal keypose, the network is encouraged to infer the final target pose directly from image observations and the high-dimensional skill embedding, thereby improving spatial understanding and yielding more precise localization of the terminal configuration of each skill.

Second, the same mechanism induces a simple and robust strategy for skill transitions at inference time. Instead of invoking the diffusion sampler at every timestep, our method amortizes its cost over an entire chunk, substantially reducing the number of diffusion calls. The policy continues generating actions until the predicted subsequent action is sufficiently close to the predicted keypose in action space, at which point $\rho_\theta$ is triggered to sample the next skill. This proximity-based criterion avoids hand-crafted termination heuristics or additional classifiers, while aligning the end of each chunk with the terminal pose of its corresponding skill.

\section{Experiments}
\label{sec:experiments}

\subsection{Experiment Setup}

\begin{figure*}[t]
    \centering
    \includegraphics[width=\linewidth]{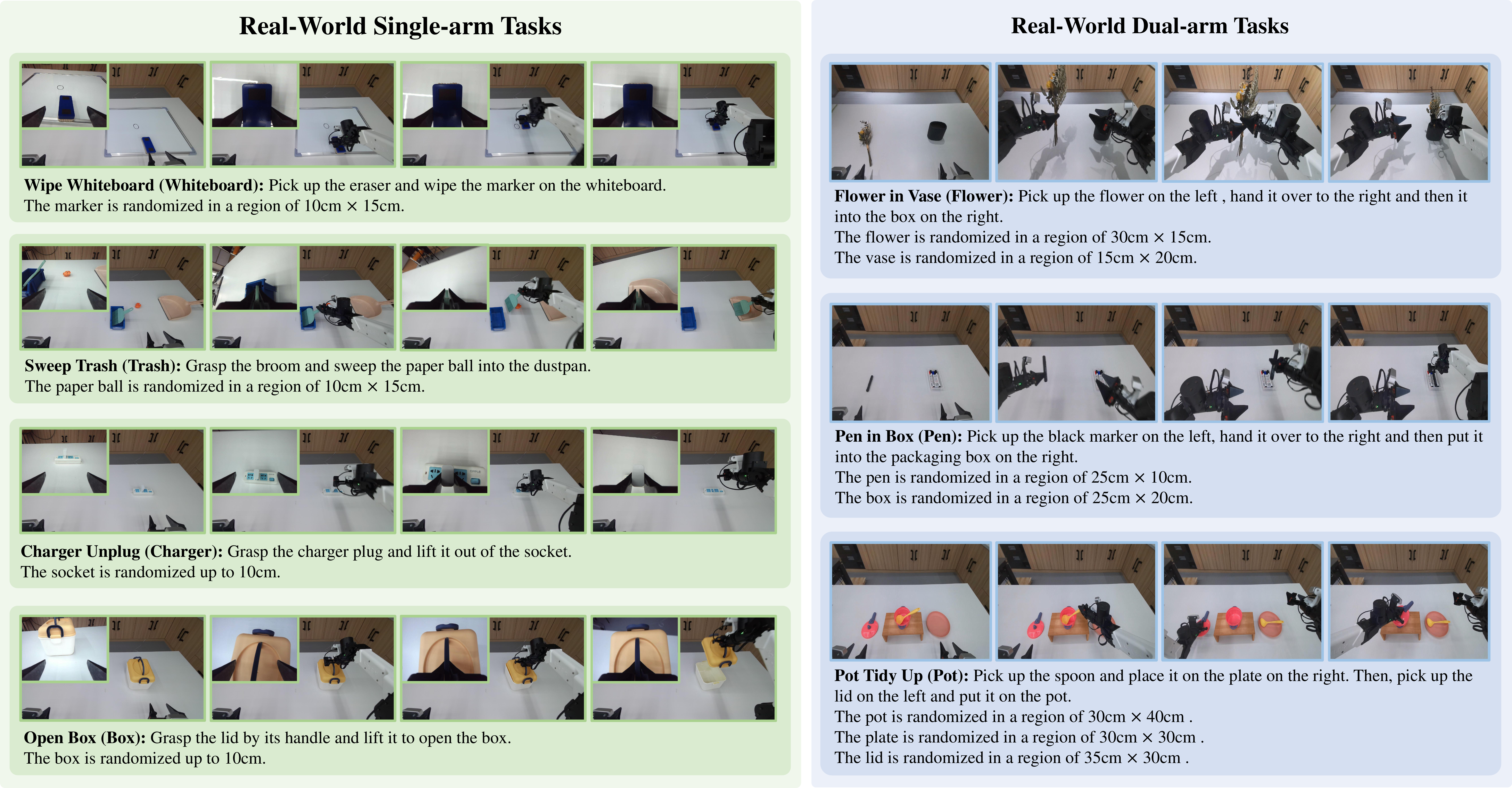}
    \caption{Illustration of tasks designed for real-world experiments. These tasks cover a broad spectrum of manipulation scenarios, ranging from short-horizon single-arm tasks with low variance to long-horizon dual-arm collaborative tasks featuring extensive spatial variability. This diverse set of tasks is designed to comprehensively test the policy's adaptability across varying levels of task complexity and workspace ranges.
    }
    \label{fig:exp illustration}
\end{figure*}

We evaluate our approach on a suite of simulated and real-world manipulation tasks.
We use RLBench~\cite{james2019rlbench} as the simulation benchmark. Compared with LIBERO~\cite{liu2023libero} and MetaWorld~\cite{yu2020meta}, which largely emphasize pick-and-place variants or short single-stage motions, RLBench provides visually rich tasks with greater structural diversity and multi-stage interactions, making it a more suitable testbed for our setting.

For real-world evaluation, we design both bimanual and single-arm tasks to assess the practicality and robustness of our method.

\begin{figure}[tb]
    \centering
    \includegraphics[width=0.8\linewidth]{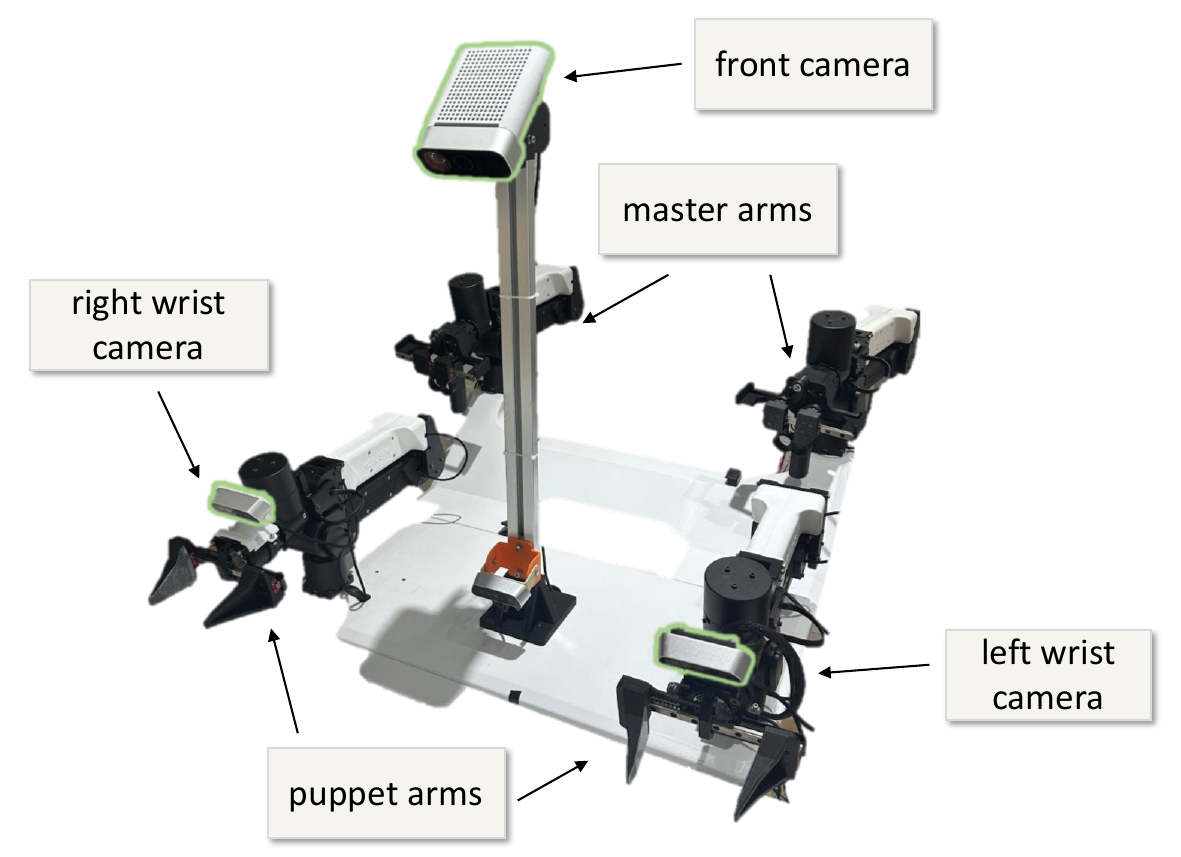}
    \caption{Real-world robot setup with dual 6-DoF puppet arms and parallel-jaw grippers, teleoperated via two kinematically matched master arms. }
    \label{fig:hardware}
\end{figure}

RLBench uses a Franka Panda arm with front-view and wrist-view RGB cameras as observations.
In the real world, we adopt an ALOHA-style~\cite{zhao2023learning, fu2024mobile} dual-arm robot equipped with one front camera and two wrist cameras (see \cref{fig:hardware}), and use their RGB images as input.
All images in both settings are resized to a resolution of $224\times224$.

\noindent\textbf{Expert Demonstrations.}
In RLBench, we select twelve most complex tasks and collect 100 demonstrations per task using the provided script-based policies.
In the real-world setting, we design three bimanual tasks and four single-arm tasks, and teleoperate the robot to collect 100 and 50 demonstrations for each task, respectively.
Illustrations of tasks in real-world experiments are provided in \cref{fig:exp illustration}.

\noindent\textbf{Baseline Methods.}
We compare with discrete LVM baselines, strong IL methods, and a recent VLA model:

\begin{itemize}
    \item \textbf{Diffusion Policy (DP)}~\cite{chi2023diffusion}: diffusion-based action generation conditioned on observations.
    \item \textbf{ACT}~\cite{zhao2023learning}: a conditional VAE policy over actions given visual and proprioceptive inputs.
    \item \textbf{MT-ACT}~\cite{bharadhwaj2024roboagent}: ACT with FiLM-conditioned language for multi-task learning.
    \item \textbf{QueST}~\cite{mete2024quest}: discrete actions via FSQ~\cite{mentzer2023finite} with causal inductive bias.
    \item \textbf{RDT}~\cite{liu2024rdt}: a state-of-the-art bimanual VLA method with a 1B-parameter diffusion foundation model pretrained on large-scale robot datasets.
\end{itemize}
\noindent\textbf{Metrics.}
We report both \textit{Success Rate (SR)} and \textit{Average Task Progress (ATP)}.
SR is a standard binary metric that only returns $100\%$ when all success conditions are met, but it often fails to reflect meaningful partial completion in long-horizon tasks.
To address this limitation, we introduce ATP to provide a finer-grained view of task performance.
We decompose each task into a sequence of segments (\cref{subsec: skill discovery}) and compute ATP as the normalized proportion of completed segments, yielding a value in $[0,1]$.
We use SR for simulation tasks, where success is reliably detected, and report both SR and ATP in real-world experiments for a more comprehensive evaluation.

\noindent\textbf{Evaluation Settings.}
Previous methods such as ACT~\cite{zhao2023learning} and DP~\cite{chi2023diffusion} train separate policies per task, avoiding inter-task interference but sidestepping key challenges of multi-task imitation learning—namely, handling behavioral multi-modality, noisy trajectories, and distribution shifts under a shared policy. 
To directly evaluate these challenges, we train and test our method in a multi-task setting, and include single-task results for ACT and DP for comparison.

\begin{table*}[ht]
\caption{Simulation performance across twelve tasks measured by SR (\%). An asterisk (*) indicates single-task training; all other methods use the multi-task setting.}
\label{tab:rlbench main}
\resizebox{1\linewidth}{!}{
\begin{tabular}{lccccccccccccc}
\toprule
Methods              & Average                                           & Open Box & Close Box & Dustpan & Laptop & Toilet Up & Toilet Down & Rubbish & Phone & Umbrella & Jenga & Lid   & Scales \\ \midrule
DP$^*$               & \cellcolor[HTML]{EFEFEF}42.8 & 16.7     & \textbf{86.7}      & 66.7    & \textbf{90.0}   & 46.7      & \underline{96.7}        & 6.7     & 10.0  & 0.0      & 40.0  & 43.3  & 10.0   \\
ACT$^*$              & \cellcolor[HTML]{EFEFEF}46.9 & 46.7     & 66.7      & 73.3    & \underline{76.7}   & 13.3      & \underline{96.7}        & 26.7    & 13.3  & \textbf{26.7}     & 66.7  & 56.7  & 0.0    \\ \midrule
DP          & \cellcolor[HTML]{EFEFEF}36.9  & 10.0 & 63.3 & 40.0 & \underline{76.7}& 20.0 & \underline{96.7} & 6.7 & 16.7 & 13.3 & 30.0 & 70.0 & 0.0       \\ 
MT-ACT               & \cellcolor[HTML]{EFEFEF}\underline{68.6} & \underline{50.0}     & 63.3      & \textbf{93.3}    & 63.3   & \textbf{83.3}      & \textbf{100.0}       & \underline{53.3}    & \underline{63.3}  & 16.7     & \textbf{90.0}  & \textbf{100.0} & \underline{46.7}   \\
QueST                & \cellcolor[HTML]{EFEFEF}32.8 & 0.0      & 63.3      & 53.3    & 43.3   & 13.3      & 73.3        & 23.3    & 20.0  & 16.7     & 26.7  & 50.0  & 10.0   \\
\ours & \cellcolor[HTML]{EFEFEF}\textbf{74.2} & \textbf{80.0}     & \underline{80.0}      & \underline{90.0}    & 66.7   & \underline{76.7}      & \textbf{100.0}       & \textbf{56.7}    & \textbf{73.3}  & \underline{20.0}     & \underline{86.7}  & \underline{83.3}  & \textbf{76.7}   \\ \bottomrule
\end{tabular}
}
\end{table*}

\subsection{Simulation Results}
\label{subsec:sim_exp}
We select twelve RLBench tasks designed to assess skill transfer and generalization under a multi-task setting, spanning diverse manipulation primitives and task structures.

The twelve tasks are chosen to evaluate two key capabilities: the ability to reproduce consistent motion dynamics (e.g., smooth closing in \textit{Close Box}), and the ability to perform accurate spatial localization, including multi-stage localization (e.g., jointly identifying the rubbish and the bin in \textit{Rubbish in Bin}, or reasoning about the umbrella pose before grasping in \textit{Umbrella out of Stand}). These tasks reflect common challenges in real-world manipulation, including trajectory consistency and spatial reasoning. 

We compare our approach against both single-task and multi-task baselines.
While single-task methods such as DP~\cite{chi2023diffusion} and ACT~\cite{zhao2023learning} avoid inter-task interference, they fail to capture cross-task regularities and therefore exhibit limited overall performance.
Among multi-task baselines, MT-ACT~\cite{bharadhwaj2024roboagent} benefits from shared training, reaching $68.6\%$ average SR, which suggests that large-capacity transformer-based models can partially exploit common structure across tasks.
Nevertheless, MT-ACT~\cite{bharadhwaj2024roboagent} still underperforms \ours, particularly on tasks with longer horizons and higher spatial variability.
In contrast, DP~\cite{chi2023diffusion} and QueST~\cite{mete2024quest} show limited scalability in the multi-task regime, achieving only $36.9\%$ and $32.8\%$ SR, respectively, likely due to the absence of explicit skill abstraction or cross-task alignment mechanisms.


\subsection{Real-world Results}
\label{sec:realworld_exp}
We evaluate our framework on four single-arm tasks (\textit{Wipe Whiteboard}, \textit{Sweep Trash}, \textit{Charger Unplug}, and \textit{Open Box}) and three bimanual tasks (\textit{Pen in Container}, \textit{Flower in Vase}, and \textit{Pot Tidy Up}) using a dual-arm robot.

The single-arm tasks test fundamental manipulation capabilities. All methods are trained on 200 demonstrations (50 per task). As shown in \cref{tab:real_exp_single}, baselines achieve similar ATP scores (0.68–0.73), while our method substantially outperforms them with 0.93 ATP.

The bimanual tasks further stress precise localization and long-horizon coordination. \ours{} achieves the best performance on all three tasks despite being trained from scratch with only 100 demonstrations per task, while RDT~\cite{liu2024rdt} relies on large-scale pretraining (\cref{tab:real_exp_bimanual}). The keyframe-action design is particularly beneficial for spatial reasoning by explicitly predicting terminal actions, guiding accurate goal inference. For the long-horizon \textit{Pot Tidy Up} task, skill-level chunking with keyposes enables coherent dual-arm coordination, again outperforming all baselines.

\begin{table}[tb]
\centering
\caption{Real-world single-arm experiment results (ATP / SR).}
\label{tab:real_exp_single}
\resizebox{1\linewidth}{!}{
\begin{tabular}{lccccc}
\toprule
Methods & Average (ATP / SR) & Whiteboard & Charger & Box & Trash \\
\midrule
DP
& \cellcolor[HTML]{EFEFEF}0.69 / 63.0
& 0.75 / 60.0
& 0.60 / 60.0
& 0.80 / 80.0
& 0.60 / 50.0 \\

ACT
& \cellcolor[HTML]{EFEFEF}0.73 / 70.0
& \textbf{1.00} / 100.0
& 0.70 / 70.0
& 0.60 / 60.0
& 0.60 / 50.0 \\

QueST
& \cellcolor[HTML]{EFEFEF}0.69 / 63.0
& 0.75 / 60.0
& 0.70 / 70.0
& 0.60 / 60.0
& 0.70 / 60.0 \\

RDT
& \cellcolor[HTML]{EFEFEF}0.68 / 60.0
& 0.70 / 50.0
& 0.60 / 60.0
& 0.70 / 70.0
& 0.70 / 60.0 \\

\ours
& \cellcolor[HTML]{EFEFEF}\textbf{0.93 / 90.0}
& \textbf{1.00} / \textbf{100.0}
& \textbf{0.80} / \textbf{80.0}
& \textbf{1.00} / \textbf{100.0}
& \textbf{0.90} / \textbf{80.0} \\
\bottomrule
\end{tabular}}
\end{table}

\begin{table}[]
\caption{Real-World bimanual experiment results (ATP / SR).}
\label{tab:real_exp_bimanual}
\resizebox{\linewidth}{!}{
\begin{tabular}{lcccc}
\toprule
\multicolumn{1}{c}{\multirow{2}{*}{Methods}} & \multicolumn{4}{c}{ATP ($\cdot$/1) / SR ($\%$)}                                                              \\ \cline{2-5} 
\multicolumn{1}{c}{}                         & Average              & Pen                  & Flower               & Pot                  \\ \hline
DP                                          & \cellcolor[HTML]{EFEFEF}{0.18 / 0.0}           & 0.13 / 0.0           & 0.00/ 0.0            & 0.40 / 0.0           \\
MT-ACT                                          & \cellcolor[HTML]{EFEFEF}{0.34 / 13.3}          & 0.47 / 30.0          & 0.23 / 10.0          & 0.33 / 0.0           \\
QueST                                        & \cellcolor[HTML]{EFEFEF}{0.24 / 3.3}           & 0.13 / 0.0           & 0.25 / 10.0          & 0.36 / 0.0           \\
RDT                                        & \cellcolor[HTML]{EFEFEF}{0.28 / 6.7}           & 0.10 / 10.0          & 0.3 / 10.0           & 0.45 / 0.0           \\
\ours                                         & \cellcolor[HTML]{EFEFEF}{\textbf{0.60 / 46.7}} & \textbf{0.67 / 60.0} & \textbf{0.48 / 10.0} & \textbf{0.65 / 20.0} \\ \bottomrule
\end{tabular}}
\end{table}

\begin{table}[tb]
\centering
\caption{Ablation on contrastive learning terms. Both losses improve over plain VQ, and combining $\mathcal{L}_{\text{temp}}$ and $\mathcal{L}_{\text{skill}}$ achieves the highest success rate (\%), particularly on localization tasks.}
\label{tab:ablation_contrast}
\resizebox{1.0\linewidth}{!}{
\begin{tabular}{ccccc}
\toprule
\multirow{2}{*}{$\mathcal{L}_{\text{temp}}$} &
\multirow{2}{*}{$\mathcal{L}_{\text{skill}}$} &
\multicolumn{3}{c}{Success Rate (\%)} \\ \cline{3-5}
 & & Average $\uparrow$ & Motion Pattern Tasks & Localization Tasks \\ 
\midrule
           &             & \cellcolor[HTML]{EFEFEF}15.8 & 23.9 & 7.8 \\
\checkmark &             & \cellcolor[HTML]{EFEFEF}63.3 & 77.8 & 48.9 \\
           & \checkmark  & \cellcolor[HTML]{EFEFEF}70.0 & 81.1 & 58.9 \\
\checkmark & \checkmark  & \cellcolor[HTML]{EFEFEF}\textbf{74.2} & \textbf{82.2} & \textbf{66.1} \\
\bottomrule
\end{tabular}
}
\end{table}

\begin{figure}[h]
    \centering
    \includegraphics[width=\linewidth]{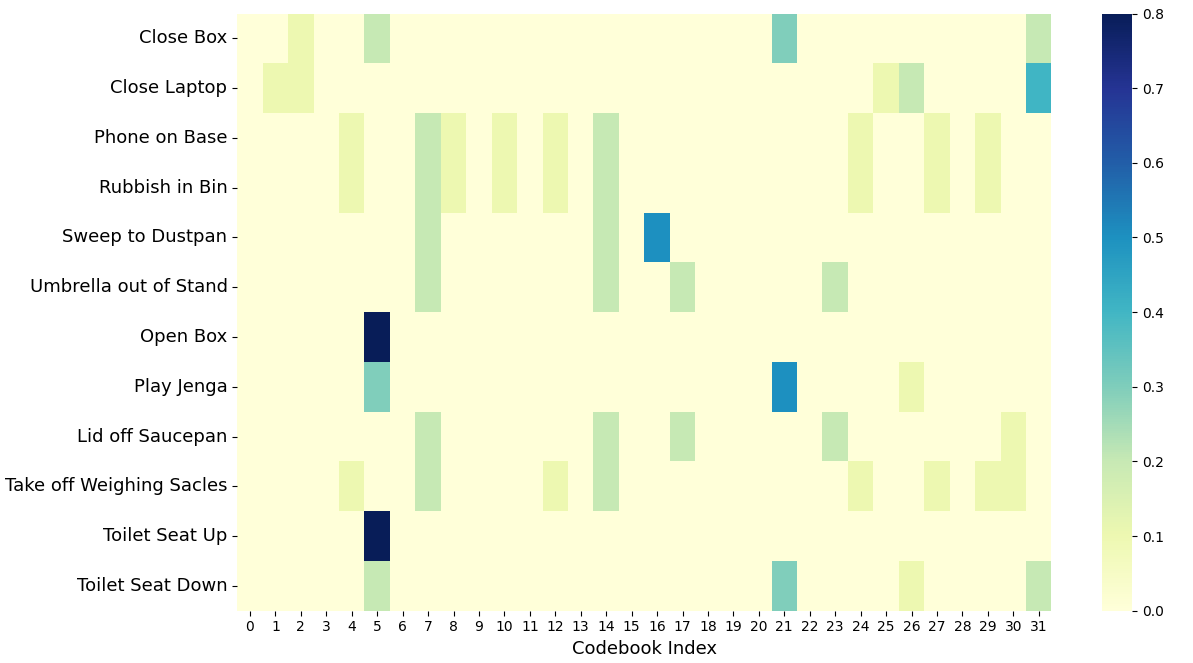}
    \caption{Visualization of skill heat map on twelve RLBench tasks.}
    \label{fig:skill_vis}
\end{figure}

\subsection{Ablation Study}
To analyze the contribution of each component, we conduct ablation studies on twelve RLBench tasks. For analysis, we divide them into \emph{Motion Pattern Tasks} (\textit{Close Box}, \textit{Open Box}, \textit{Sweep to Dustpan}, \textit{Close Laptop}, \textit{Toilet Up}, \textit{Toilet Down}) and \emph{Spatial Localization Tasks} (\textit{Rubbish in Bin}, \textit{Phone on Base}, \textit{Umbrella out of Stand}, \textit{Play Jenga}, \textit{Take off Weighing Scales}, \textit{Lid off Saucepan}), reflecting differences between execution regularity and spatial reasoning.

\noindent \textbf{Contrastive Objectives.}
As shown in \cref{tab:ablation_contrast}, both $\mathcal{L}_{\text{temp}}$ and $\mathcal{L}_{\text{skill}}$ are crucial for structured skill learning. Removing either degrades performance, particularly on localization tasks. While $\mathcal{L}_{\text{temp}}$ enforces temporal consistency within skill segments, $\mathcal{L}_{\text{skill}}$ promotes semantic separation across skills. Using both yields the best results, indicating that temporal coherence and semantic discrimination are complementary for learning a transferable skill space.

\noindent \textbf{Keyframe Action Prediction.}
As shown in \cref{tab:ablation_keypose}, predicting the next keyframe action consistently improves performance, increasing the average success rate from 73.3\% to 74.2\%. Gains mainly come from localization tasks, whose success rate rises from 61.1\% to 66.1\%, while motion-pattern tasks remain largely unchanged. This suggests that keyframe prediction improves spatial grounding and goal inference, enabling more accurate multi-step planning without affecting smooth motion execution.

\begin{table}[tb]
\centering
\caption{Ablation on keyframe action prediction. Predicting the next keyframe improves success rate ($73.3\%\!\rightarrow\!74.2\%$), mainly benefiting localization tasks.}
\label{tab:ablation_keypose}
\resizebox{1.0\linewidth}{!}{%
\begin{tabular}{cccc}
\toprule
\multirow{2}{*}{\shortstack{Keyframe\\Action}} & \multicolumn{3}{c}{Success Rate (\%)} \\ \cline{2-4}
 & Average $\uparrow$ & Motion Pattern Tasks & Localization Tasks \\
\midrule
            & \cellcolor[HTML]{EFEFEF}73.3 & \textbf{85.5} & 61.1 \\
\checkmark  & \cellcolor[HTML]{EFEFEF}\textbf{74.2} & 82.2 & \textbf{66.1} \\
\bottomrule
\end{tabular}
}
\end{table}

\begin{figure}[h]
    \centering
    \includegraphics[width=0.9\linewidth]{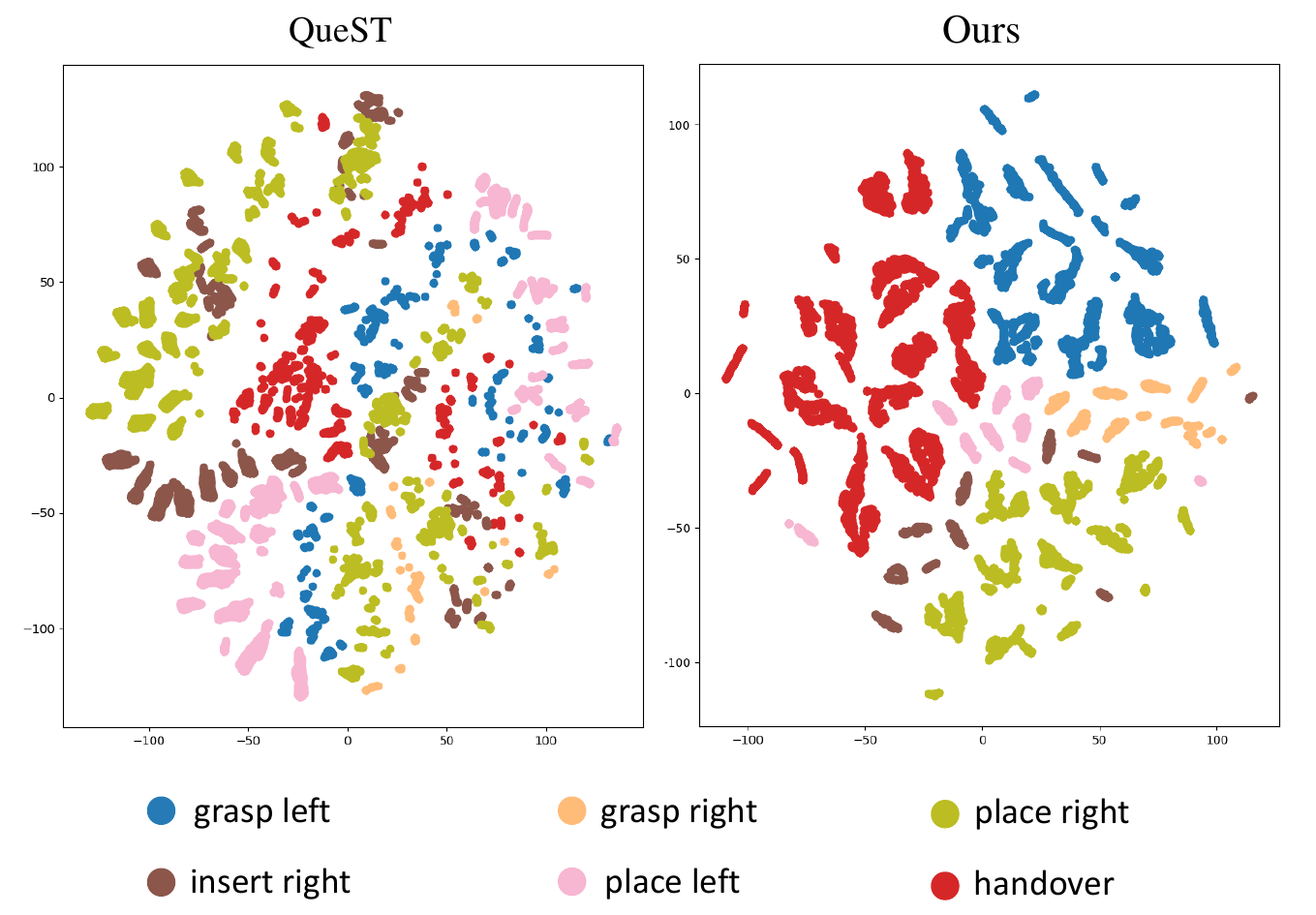}
    \caption{t-SNE visualization of latent features of QueST and \ours\ in the real-world bimanual task setting.}
    \label{fig:tsne}
\end{figure}

\subsection{Visualization of Learned Atomic Skills}

We visualize the normalized atomic skill usage across twelve RLBench tasks as a heatmap, where rows denote tasks and columns denote skill codes. Clear cross-task reuse patterns emerge. For example, skills 7 and 14 are consistently activated in grasping-related tasks (e.g., Open Box, Play Jenga, Toilet Seat Up) but rarely appear in object-contact tasks such as Rubbish in Bin or Sweep to Dustpan. Overall, skill usage is sparse and task-specific, indicating that the learned atomic skill space captures semantically coherent and selectively reusable sub-behaviors.

\subsection{Visualization of the Atomic Skill Library}
We present a t-SNE visualization of the learned atomic skill library in the real-world bimanual setting. As shown in Fig. 5, QueST [12] focuses on low-level motion patterns without modeling high-level skills, resulting in fragmented and overlapping feature clusters. In contrast, our framework learns a semantically structured atomic skill library at a higher level of abstraction, which better supports multi-task robot manipulation.

\section{Conclusion}
We presented AtomSkill, a multi-task imitation learning framework that learns a structured atomic skill library for cross-task generalization. By combining semantic skill alignment, keypose imagination, and diffusion-based sampling, AtomSkill enables composable and transferable robot manipulation. Experiments in both simulation and real-world settings demonstrate consistent improvements over existing methods.
A limitation of our approach lies in defining atomic skill granularity for continuous tasks without clear transitions (e.g., spreading sauce), which may introduce ambiguity in the learned skill library.




\label{sec:conclusion}

\section*{Acknowledgement}
This work was supported by NSFC (No.62406195, 62303319), Shanghai Frontiers Science Center of Human-centered Artificial Intelligence (ShangHAI), MoE Key Laboratory of Intelligent Perception and Human-Machine Collaboration (ShanghaiTech University), the Shanghai Frontiers Science Center of Human-centered Artificial Intelligence, HPC Platform and Core Facility Platform of Computer Science and Communication of ShanghaiTech University and Shanghai Engineering Research Center of Intelligent Vision and Imaging.

\bibliographystyle{IEEEtran}
\bibliography{IEEEabrv,reference}

\newpage
\section*{APPENDIX}

\section{Overview}

This supplementary document provides additional implementation details, experimental results, and qualitative analyses that complement the main paper. Specifically, we include:

\begin{itemize}
    \item Detailed descriptions of the simulation and real-world experimental setups;
    \item Additional discussions of the performance of baseline methods;
    \item A discussion of the limitations of the proposed \ours{} framework.
\end{itemize}

\section{Experiment Details}

In this section, we provide additional details of both the simulation environment and the real-world experimental setup.

\subsection{Simulation Overview}

\begin{figure*}[t]
    \centering
    \includegraphics[width=0.9\linewidth]{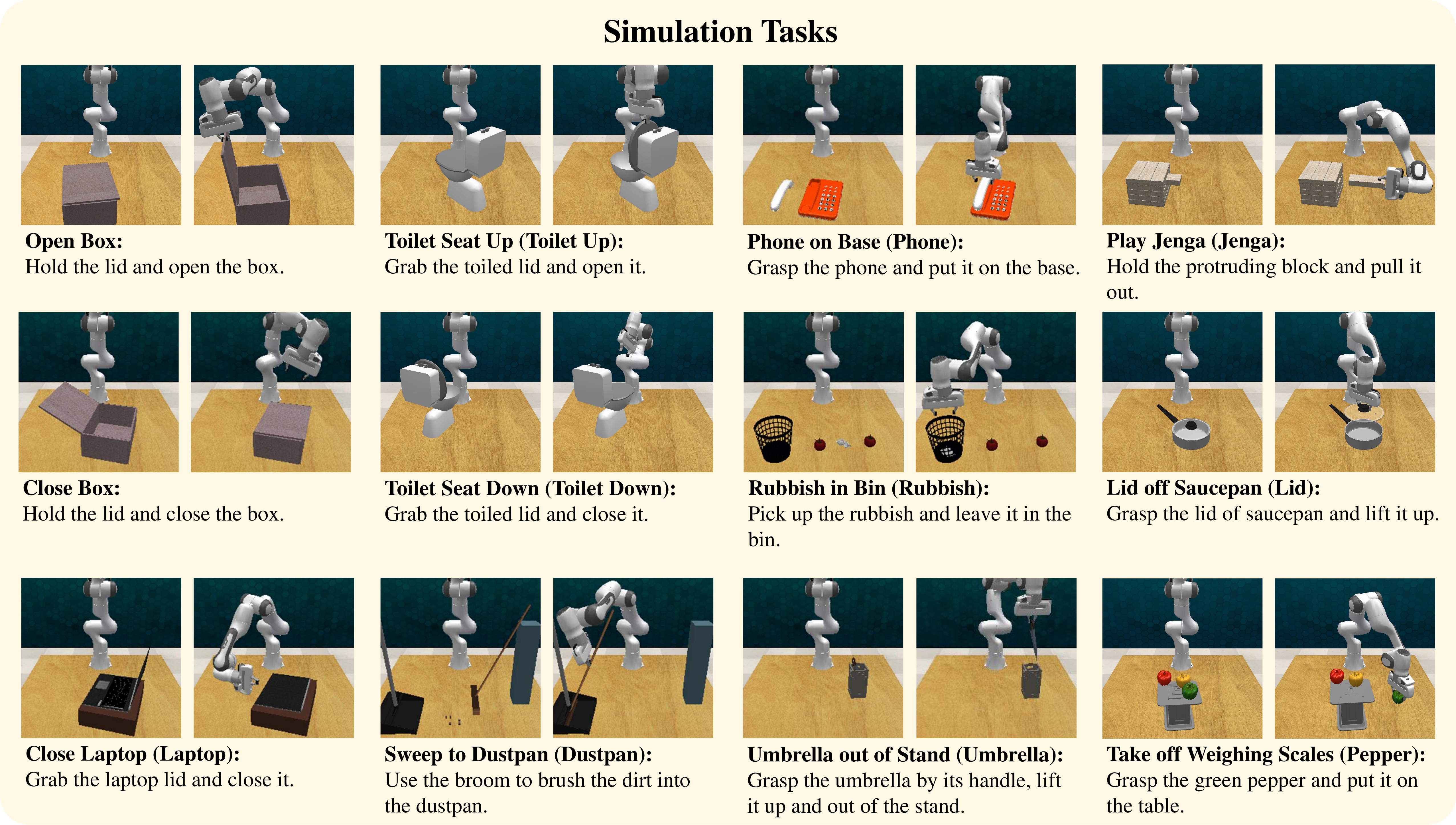}
    \caption{
    Illustration of the twelve RLBench manipulation tasks used in simulation. Each task can be decomposed into a small number of reusable atomic skills, enabling the evaluation of cross-task skill sharing and generalization.
    }
    \label{fig:sim_exp_illustration}
\end{figure*}

We evaluate our method on twelve tabletop manipulation tasks from RLBench. These tasks span both object manipulation and environment interaction, as illustrated in~\cref{fig:sim_exp_illustration}. Each task naturally decomposes into two or three reusable manipulation skills, while multiple tasks share common skill primitives. Such characteristics make RLBench a suitable benchmark for evaluating the quality, compositionality, and transferability of learned atomic skills.

\subsection{Real-World Hardware Setup}

We conduct all real-world experiments on an ALOHA-style dual-arm robotic platform~\cite{zhao2023learning,fu2024mobile}, as shown in~\cref{fig:hardware}. The platform consists of two 6-DoF puppet arms equipped with parallel-jaw grippers for manipulation and two master arms with identical kinematic structures for teleoperation. Visual observations are provided by three RGB cameras, including two wrist-mounted cameras and one overhead camera positioned between the two robot arms. All experiments are performed on a workstation equipped with an NVIDIA RTX 4090 GPU (24 GB).

\subsection{Real-World Experiment Task Details}
\label{sec:real_detail}

We evaluate our method on seven real-world manipulation tasks, including three bimanual tasks and four single-arm tasks. Table~\ref{tab:task_summary} summarizes each task together with its average trajectory length, the number of demonstrations, and the corresponding language instruction.

To provide a more fine-grained evaluation than binary success rates, we adopt the \textbf{Average Task Progress (ATP)} metric. Each task is decomposed into several sequential stages, and the ATP score is computed by normalizing the achieved stage index. Detailed task progress definitions for both bimanual and single-arm tasks are presented in Tables~\ref{tab:real_dual} and~\ref{tab:real_single}, respectively.

\begin{table*}[t]
\centering
\caption{
Summary of the real-world manipulation tasks used in our experiments, including the average trajectory length, the number of demonstrations, and the corresponding language instruction.
}
\label{tab:task_summary}

\resizebox{0.76\linewidth}{!}{
\begin{tabular}{cccc}
\toprule

Task &
\begin{tabular}[c]{@{}c@{}}
Average\\
Trajectory Length
\end{tabular}
&
\begin{tabular}[c]{@{}c@{}}
Number of\\
Demonstrations
\end{tabular}
&
Task Instruction
\\

\midrule

\multicolumn{4}{c}{Real-World Bimanual Tasks}\\
\midrule

Flower in Vase
&
338.86
&
100
&
Pick up the flower and insert it into the vase.
\\

Pen in Container
&
328.25
&
100
&
Pick up the pen and place it into the container.
\\

Pot Tidy Up
&
427.89
&
100
&
\begin{tabular}[c]{@{}c@{}}
Pick up the spoon from the pot, place it onto the plate,\\
and finally cover the pot with its lid.
\end{tabular}
\\

\midrule

\multicolumn{4}{c}{Real-World Single-Arm Tasks}\\
\midrule

Wipe Whiteboard
&
180.44
&
50
&
Pick up the eraser and wipe the whiteboard.
\\

Sweep Trash
&
190.17
&
50
&
Pick up the broom and sweep the trash into the dustpan.
\\

Charger Unplug
&
148.33
&
50
&
Grasp the charger and unplug it from the socket.
\\

Open Box
&
155.33
&
50
&
Grasp the lid handle and open the box.
\\

\bottomrule
\end{tabular}
}
\end{table*}

\noindent\textbf{Bimanual Tasks.}
We design three representative bimanual manipulation tasks on the ALOHA platform, namely \emph{Flower in Vase}, \emph{Pen in Container}, and \emph{Pot Tidy Up}. To provide a more informative evaluation than binary success rates, we adopt the \textbf{Average Task Progress (ATP)} metric. Specifically, each task is decomposed into a sequence of ordered stages, and a trial is considered to have reached the highest successfully completed stage before failure. The ATP score is obtained by normalizing the achieved stage index, enabling fair comparisons across tasks with different horizons. During data collection and evaluation, all manipulated objects are randomly initialized within a $10\,\mathrm{cm}\times10\,\mathrm{cm}$ workspace to introduce spatial variations.

The detailed stage definitions are summarized in Table~\ref{tab:real_dual}. Visual illustrations of the three tasks are provided in~\cref{fig:exp illustration}.

\begin{table*}[t]
\centering
\caption{
Definitions of task progress for the real-world bimanual tasks. We report the unnormalized task progress for clarity. Each task is decomposed into multiple sequential stages, and failure at any stage terminates the evaluation. The reported Average Task Progress (ATP) is obtained by normalizing these stage scores.
}
\label{tab:real_dual}

\begin{tabular}{ccl}
\toprule

Task &
\begin{tabular}[c]{@{}c@{}}
Unnormalized\\
Task Progress
\end{tabular}
&
Stage
\\

\midrule

\multirow{3}{*}{Flower in Vase}

&1&
Pick up the flower by grasping the stem with the left gripper; failure is declared if the flower head is grasped.
\\

&2&
Transfer the flower to the right gripper; failure is declared if the flower head is grasped during handover.
\\

&3&
Insert the flower into the vase; failure is declared if the flower drops onto the table.
\\

\midrule

\multirow{3}{*}{Pen in Container}

&1&
Pick up the pen with the left gripper.
\\

&2&
Transfer the pen to the right gripper.
\\

&3&
Place the pen into the container; failure is declared if the pen falls onto the table.
\\

\midrule

\multirow{4}{*}{Pot Tidy Up}

&1&
Pick up the spoon from the pot using the right gripper.
\\

&2&
Place the spoon onto the plate.
\\

&3&
Grasp the pot lid with the left gripper.
\\

&4&
Cover the pot with the lid; failure is declared if the lid slides off the pot.
\\

\bottomrule
\end{tabular}

\end{table*}

\vspace{0.5em}

\noindent\textbf{Single-Arm Tasks.}
To further evaluate policy performance under simpler manipulation settings, we additionally design four single-arm tasks on the same robotic platform, where only the right arm is activated. For these experiments, visual observations consist only of the front-view camera and the wrist-mounted camera on the active arm. Similar to the bimanual setting, each task is decomposed into multiple sequential stages for ATP evaluation. The complete task progress definitions are presented in Table~\ref{tab:real_single}, while qualitative examples are shown in~\cref{fig:exp illustration}.

\begin{table*}[t]
\centering
\caption{
Definitions of task progress for the real-world single-arm tasks. Similar to the bimanual setting, the reported Average Task Progress (ATP) is computed by normalizing the corresponding stage scores.
}
\label{tab:real_single}

\begin{tabular}{ccl}
\toprule

Task &
\begin{tabular}[c]{@{}c@{}}
Unnormalized\\
Task Progress
\end{tabular}
&
Stage
\\

\midrule

\multirow{2}{*}{Wipe Whiteboard}

&1&
Grasp the eraser near its center.
\\

&2&
Erase the marker on the whiteboard; failure is declared if the marker is not completely removed.
\\

\midrule

\multirow{2}{*}{Sweep Trash}

&1&
Grasp the broom handle.
\\

&2&
Sweep the paper ball into the dustpan.
\\

\midrule

\multirow{2}{*}{Charger Unplug}

&1&
Grasp the charger plug.
\\

&2&
Pull the charger out of the socket.
\\

\midrule

\multirow{2}{*}{Open Box}

&1&
Grasp the lid handle.
\\

&2&
Lift the lid to open the box.
\\

\bottomrule
\end{tabular}

\end{table*}

\subsection{Training}

Our framework consists of three main components: the skill encoder $\phi_\theta$, the skill diffusion sampler $\rho_\theta$, and the action decoder $\psi_\theta$. The skill encoder comprises several convolutional layers followed by a six-layer self-attention transformer for learning compact atomic skill representations. The skill diffusion sampler is implemented using a CNN-based diffusion policy~\cite{chi2023diffusion} with FiLM conditioning to generate skill sequences conditioned on visual observations. Finally, the action decoder consists of two cross-attention transformer blocks with seven layers each, which translate predicted skill sequences into executable robot actions.

The default hyperparameters used throughout all experiments are summarized in Table~\ref{suppl:Policy}.

\begin{table}[t]
\centering
\caption{
Default hyperparameters used for training the proposed policy.
}
\label{suppl:Policy}

\resizebox{0.65\linewidth}{!}{
\begin{tabular}{lc}
\toprule
Hyperparameter & Value\\
\midrule
Training epochs ($\phi_\theta,\psi_\theta$) & 200\\
Training epochs ($\rho_\theta$) & 1000\\
Batch size & 256\\
Action horizon & 32\\
Codebook size & 32\\
Number of skill embedding tokens & 8\\
Number of attention heads & 8\\
Attention layers ($\phi_\theta$) & 6\\
Attention layers ($\psi_\theta$) & 7\\
Learning rate & $1\times10^{-4}$\\
Weight decay & $1\times10^{-5}$\\
Commitment weight $\lambda$ & 0.25\\
Temperature $\mathcal{T}$ & 0.1\\
Reconstruction loss weight $\beta_1$ & 1\\
Contrastive loss weight $\beta_2$ & $1\times10^{-2}$\\
\bottomrule
\end{tabular}
}
\end{table}

\section{Further Discussions}

\subsection{Discussion of Latent Variable Models}

On RLBench, the discrete latent variable model QueST~\cite{mete2024quest} performs noticeably worse than ACT~\cite{zhao2023learning} and Diffusion Policy (DP)~\cite{chi2023diffusion}, despite demonstrating strong performance on several large-scale benchmarks. We attribute this discrepancy to the characteristics of the learned latent representations. QueST primarily compresses trajectories into motion-centric discrete codes optimized for action reconstruction, which are less effective at capturing semantically meaningful manipulation primitives. Consequently, its latent space provides limited support for precise spatial reasoning and goal-oriented behavior, both of which are essential for complex manipulation tasks in RLBench.

A similar trend is observed in our real-world experiments. While QueST achieves competitive performance on relatively simple single-arm tasks, its performance degrades substantially on long-horizon bimanual manipulation, indicating limited skill diversity and insufficient robustness under large spatial variations. In contrast, \ours{} explicitly learns semantically aligned atomic skills together with their corresponding terminal actions, producing a reusable and transferable skill space while preserving the spatial precision required for accurate robot control.

\subsection{Discussion of Vision-Language-Action Models}

We select RDT~\cite{liu2024rdt} as the representative Vision-Language-Action (VLA) baseline because its pretraining data closely matches our experimental setup. Besides Internet-scale robot datasets, RDT is additionally pretrained on more than 6,000 demonstrations collected using the same ALOHA embodiment adopted in our experiments. Nevertheless, RDT consistently underperforms our method on both simulation and real-world tasks.

These results suggest that carefully structured skill abstractions can substantially improve data efficiency. Rather than relying solely on large-scale pretraining, explicitly modeling reusable semantic skills enables better generalization with significantly fewer task-specific demonstrations.

\section{Limitations}
Although our framework shows strong performance in both simulation and real-world multi-task manipulation, it has several limitations. First, the skill discovery pipeline relies on gripper state changes as keyframes; this works well for tasks with clear contact events but is less reliable for continuous behaviors without obvious transitions (e.g., smoothly spreading sauce). Second, the semantic grounding of skills depends on a vision–language model and prompt design, so noisy observations or imperfect descriptions may introduce label noise and add computational overhead for large-scale annotation.

\end{document}